\documentclass[10pt,twocolumn,letterpaper]{article}

\usepackage{cvpr}
\usepackage{times}
\usepackage{epsfig}
\usepackage{graphicx}
\usepackage{amsmath}
\usepackage{amssymb}
\usepackage{amsfonts}       % blackboard math symbols
\usepackage{nicefrac}       % compact symbols for 1/2, etc.
\usepackage{microtype}      % microtypography
\usepackage{xcolor}
\usepackage{grffile}
\usepackage{amsmath}
\usepackage{graphicx}
\usepackage{verbatim}
\usepackage{setspace}
\usepackage{subcaption}
\usepackage{verbatim}
\usepackage{multirow}
\usepackage{mwe}
\usepackage{verbatimbox}
\usepackage{array,booktabs}
\usepackage[inline]{enumitem}
\newcolumntype{C}{>{$\displaystyle}c<{$}}

\newcommand{\myparagraph}[1]{\vspace{0.0em}\noindent\textbf{#1}}
\newcolumntype{C}[1]{>{\centering\let\newline\\\arraybackslash\hspace{0pt}}m{#1}}

\newcommand{\mario}[1]{\textsf{\textcolor{blue}{\textbf{Mario:} \textit{#1}}}}

\usepackage[title,toc,titletoc,page]{appendix}

\usepackage[pagebackref=true,colorlinks=false]{hyperref}

\cvprfinalcopy % *** Uncomment this line for the final submission

\def\cvprPaperID{3887} % *** Enter the wacv Paper ID here

\ifcvprfinal\fi
\begin{document}

%%%%%%%%% TITLE
%\title{Long-Term On-Board Prediction of People Trajectories in Traffic Scenes with Uncertainty Estimates}
\title{Long-Term On-Board Prediction of People in Traffic Scenes under Uncertainty}

% Authors at the same institution
%\author{First Author \hspace{2cm} Second Author \\
% Authors at different institutions
\author{Apratim Bhattacharyya, Mario Fritz, Bernt Schiele  \\ 
Max Planck Institute for Informatics, Saarland Informatics Campus, Saarbr\"{u}cken, Germany \\
\texttt{\{abhattac, mfritz, schiele \}@mpi-inf.mpg.de}  }

\makeatletter
\let\@oldmaketitle\@maketitle% Store \@maketitle
\renewcommand{\@maketitle}{\@oldmaketitle% Update \@maketitle to insert...
    \renewcommand{\arraystretch}{0.2}
    \begin{tabular}{C{0.5cm}C{3.0cm}C{3.0cm}C{3.0cm}C{3.0cm}C{2.0cm} }
    &
    \shortstack{%
        \addvbuffer[-0pt]{\includegraphics[width=0.185\textwidth, height=0.075\textheight]{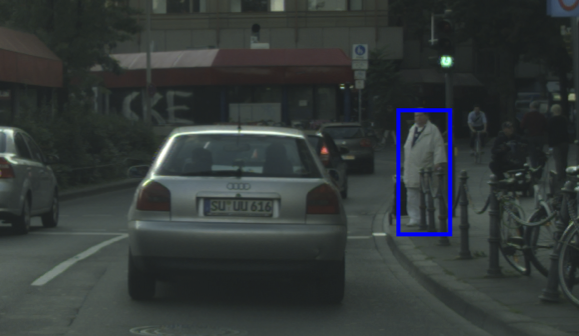}}\\
        \addvbuffer[-0pt]{\includegraphics[width=0.185\textwidth, height=0.075\textheight]{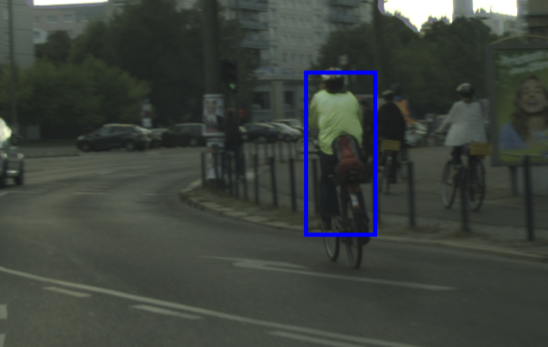}}
    }
    & 
    \shortstack{%
        \addvbuffer[-0pt]{\includegraphics[width=0.185\textwidth, height=0.075\textheight]{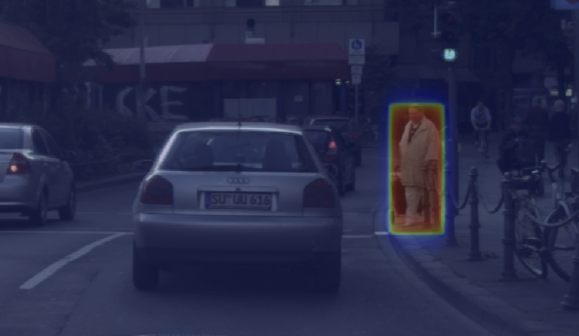}}\\
        \addvbuffer[-0pt]{\includegraphics[width=0.185\textwidth, height=0.075\textheight]{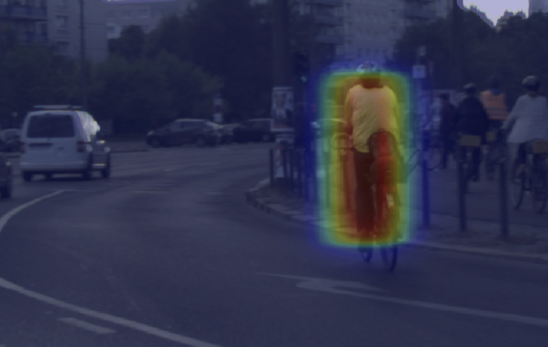}}
    }
    & 
    \shortstack{%
        \addvbuffer[-0pt]{\includegraphics[width=0.185\textwidth, height=0.075\textheight]{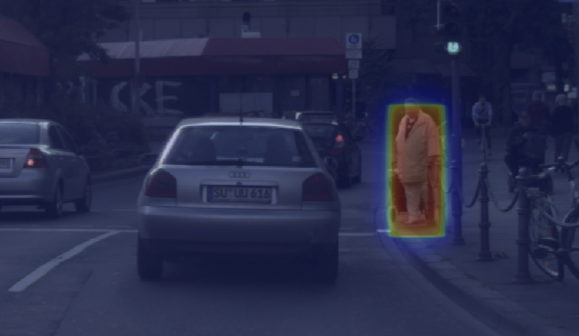}}\\
        \addvbuffer[-0pt]{\includegraphics[width=0.185\textwidth, height=0.075\textheight]{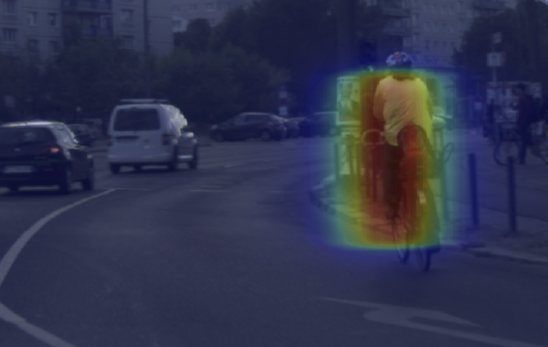}}
    }
    & 
    \shortstack{%
        \addvbuffer[-0pt]{\includegraphics[width=0.185\textwidth, height=0.075\textheight]{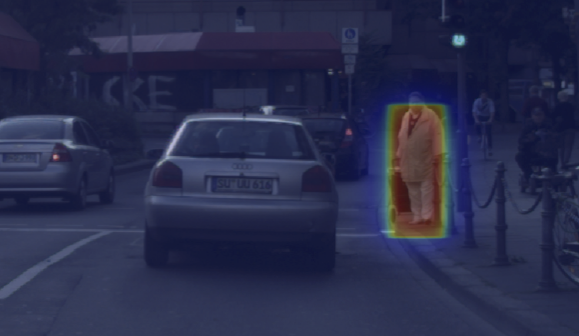}}\\
        \addvbuffer[-0pt]{\includegraphics[width=0.185\textwidth, height=0.075\textheight]{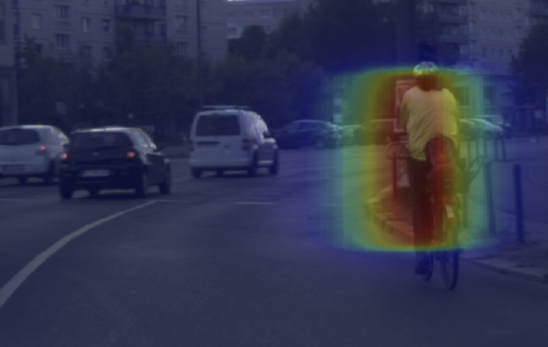}}
    }
    &
        \addvbuffer[-0pt]{\includegraphics[height=0.15\textheight]{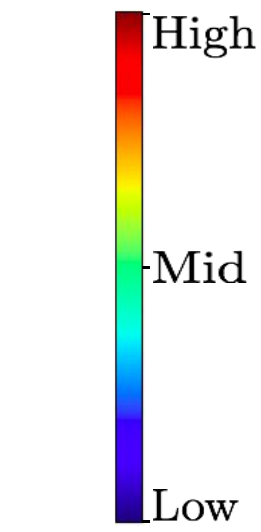}}\\

    &\textbf{Last Observation: $t$} & \textbf{Prediction: $t$ + 5} & \textbf{Prediction: $t$ + 10} & \textbf{Prediction: $t$ + 15} & \textbf{Probability} \\
    
    \\
    \end{tabular}
    \refstepcounter{figure} Figure~\thefigure: Our predictive distribution upto $t$ + 15 frames. The heat map encodes the probability of a certain pixel belonging to the person. The variance of the distribution encodes the uncertainty. \emph{Row 1}: Low uncertainty.  \emph{Row 2}: High uncertainty.
    \bigskip  \label{fig:frames}}
\makeatother

\maketitle
\ifcvprfinal\thispagestyle{empty}\fi

%%%%%%%%% ABSTRACT
\begin{abstract}
   Progress towards advanced systems for assisted and autonomous driving is leveraging  recent advances in recognition and segmentation methods. Yet, we are still facing challenges in bringing reliable driving to inner cities, as those are composed of highly dynamic scenes observed from a moving platform at considerable speeds. Anticipation becomes a key element in order to react timely and prevent accidents. 
    In this paper we argue that it is necessary to predict at least 1 second and we thus propose a new model that jointly predicts ego motion and people trajectories over such large time horizons. We pay particular attention to modeling the uncertainty of our estimates arising from the non-deterministic nature of natural traffic scenes.
    Our experimental results show that it is indeed possible to predict people trajectories at the desired time horizons and that our uncertainty estimates are informative of the prediction error. We also show that both sequence modeling of trajectories as well as our novel method of long term odometry prediction are essential for best performance.
\end{abstract}

%%%%%%%%% BODY TEXT
\section{Introduction}
While methods for automatic scene understanding have progressed rapidly over the past years, it is just one key ingredient for assisted and autonomous driving. Human capabilities go beyond inference of scene structure and encompass a broader type of scene understanding that also lends itself to anticipating the future. 

Anticipation is key in preventing collisions by predicting future movements of dynamic agents e.g. people and cars in inner cities. It is also the key to operating at practical safety distances. Without anticipation, domain knowledge and experience, drivers would have to maintain an equally large safety distance to all objects, which is clearly impractical in dense and cluttered inner city traffic. Additionally, anticipation enables decision making, e.g. passing cars and pedestrians while respecting the safety of all participants. Even at conservative and careful driving speeds of $25   {\text{miles}}/{\text{hour}}$ ($\sim 40 {\text{km}}/{\text{hour}}$) in residential areas, the distance traveled in 1 second corresponds roughly to the breaking distance. Anticipation of traffic scenes on a time horizons of \emph{at least} 1 second would therefore enable safe driving at such speeds. 

%Anticipation is possible, as the involved agents act typically in compliance with their individual goals and constraints of the scene structure. While the behavior of agents in traffic scenes is far from random, significant uncertainty in future states can accumulate due to the non-deterministic progression of the scene.

%In safety critical systems like assisted and autonomous driving, it is crucial to model uncertainty \cite{rausand2014reliability}. 

We propose the first approach to predict people (pedestrians including cyclists) trajectories from on-board cameras over such long-time horizons with uncertainty estimates. Due to the particular importance for safety, we are focusing on the people class. While pedestrian trajectory prediction has been approached in prior work, we propose the first approach for on-board prediction. As predictions are made with respect to the moving vehicle, we formulate a novel two stream model for long-term person bounding box prediction and vehicle ego motion (odometry).  In contrast to prior work, we model both \emph{aleatoric} (observation) uncertainty and \emph{epistemic} (model) uncertainty \cite{der2009aleatory} in order to arrive at an estimate of the overall uncertainty.

Our contributions in detail are:
\begin{enumerate*}
    \item First approach to long-term prediction of pedestrian bounding box sequences from a mobile platform;
    \item Novel sequence to sequence model which provides a theoretically grounded approach to quantify uncertainty associated with each prediction;
    \item Detailed experimental evaluation of alternative architectures illustrating the importance and effectiveness of using a two-stream architecture;
    \item Analysis of dependencies between uncertainty estimates and actual prediction error leading to an \emph{empirical error bound}.
\end{enumerate*}

%----------------------------------------\---------------------------------
%\mario{skipping over related work ....}
\section{Related work}
%\noindent
%{\bf Human trajectory prediction.}
\myparagraph{Human Trajectory Prediction.}
Recent works such as \cite{keller2011will,rehder2015goal} focus on the task of pedestrian trajectory prediction in 3D space.  However, 3D world coordinates are difficult to obtain in unconstrained scenarios. It requires expensive stereo camera and/or
LIDAR setups and obtained depth maps are typically noisy especially in unknown environments. Our method does not depend upon unreliable 3D coordinates and it is widely applicable as it requires only one camera. Many vehicles worldwide already have installed dash-cams. Another class of models such as \cite{helbing1995social,yamaguchi2011you,robicquet2016learning,alahi2016social,lee2017desire} consider the problem of (2D) pedestrian trajectory prediction in a social context by modelling human-human interactions. The state of the art model \cite{alahi2016social} proposes to estimate the trajectories of each person in the scene by an instance of a ``Social'' LSTM. The instances of the Social LSTM can communicate with a special pooling layer. This enables the modelling of interactions and joint estimation of trajectories of all pedestrians in the scene. In \cite{trautman2013robot} the joint estimation of robot and human trajectories are considered in a social context. However, in case of on-board prediction vehicle ego-motion dominates social aspects. Moreover, most methods are trained/tested on static camera datasets which are hand annotated with minimum observation noise. Apart from these, the class of models such as \cite{hu2007semantic,kim2011gaussian,morris2011trajectory,zhou2011random,zhang2013understanding} aim at discovering motion patterns of humans and vehicles. Such methods cannot be used for trajectory prediction and do not consider vehicle ego-motion.

\myparagraph{Modeling Uncertainty in Deep Learning.} 
Popular deep learning architectures do not model uncertainty. They assume uniform constant observation noise (aleatoric uncertainty). Heteroscedastic regression methods \cite{nix1994estimating,le2005heteroscedastic}  estimate aleatoric uncertanity by predicting the parameters of a assumed observation noise distribution  (also in \cite{alahi2016social}). Bayesian neural networks \cite{mackay1992practical,neal2012bayesian} offer a probabilistic view of deep learning and provide model (epistemic) uncertainty estimates. However, inference of model posterior in such networks is difficult. Variational Inference is a popular method. Gal et. al. in \cite{gal2016dropout} showed that dropout training in deep neural networks approximates Bayesian inference in deep Gaussian processes. Extending these results it was shown in \cite{Gal2016Bayesian} that dropout training can be cast as approximate Bernoulli variational inference in Bayesian neural networks. These results were extended to RNNs in \cite{gal2016theoretically}. The developed Bayesian RNNs showed superior performance to standard RNNs with dropout in various tasks. More recently, \cite{kendall2017uncertainties} presents a Bayesian deep learning framework jointly estimating aleatoric uncertainty together with epistemic uncertainty. The resulting framework gives new state-of-the-art results on segmentation and depth regression benchmarks.

\myparagraph{Assisted and Autonomous driving.}
One of the earliest works on vehicle ego-motion (odometry) prediction or popularly, autonomous driving, was ALVINN by \cite{pomerleau1989alvinn}. This work showed the possibility of directly predicting steering angles from visual input. This system used a simple fully-connected network. More recently, \cite{bojarski2016end} uses a convolutional neural network for this task and achieves a autonomy of 90\% using a relatively small training set. However, the focus is on highway driving. \cite{xu2016end} proposes a FCN-LSTM that predicts the next vehicle odometry based on the visual input captured by an on-board camera and previous odometry of the vehicle. Here, a diverse crowed sourced dataset is used. However, these methods predict vehicle odometry (e.g. steering angle) only for the next time-step. In contrast, we focus on inner-city driving and predict multiple time-steps into the future. \cite{santana2016learning} proposes a driving simulator that predicts the future in form of frames but suffers from blurriness problems in the long-term important details get lost. In \cite{luc2017predicting} future segmentation masks are predicted, but only mid-term (upto 0.5sec) future is predicted and there is no pedestrian specific evaluation. We predict the future in terms of bounding box coordinates which remain well defined by design in the long-term.

\section{On-board Pedestrian Prediction under Uncertainty}

\begin{figure*}[!t]
    \centering
    \begin{subfigure}{0.6\textwidth}
        \centering
        \includegraphics[width=\textwidth, height = 7cm, keepaspectratio]{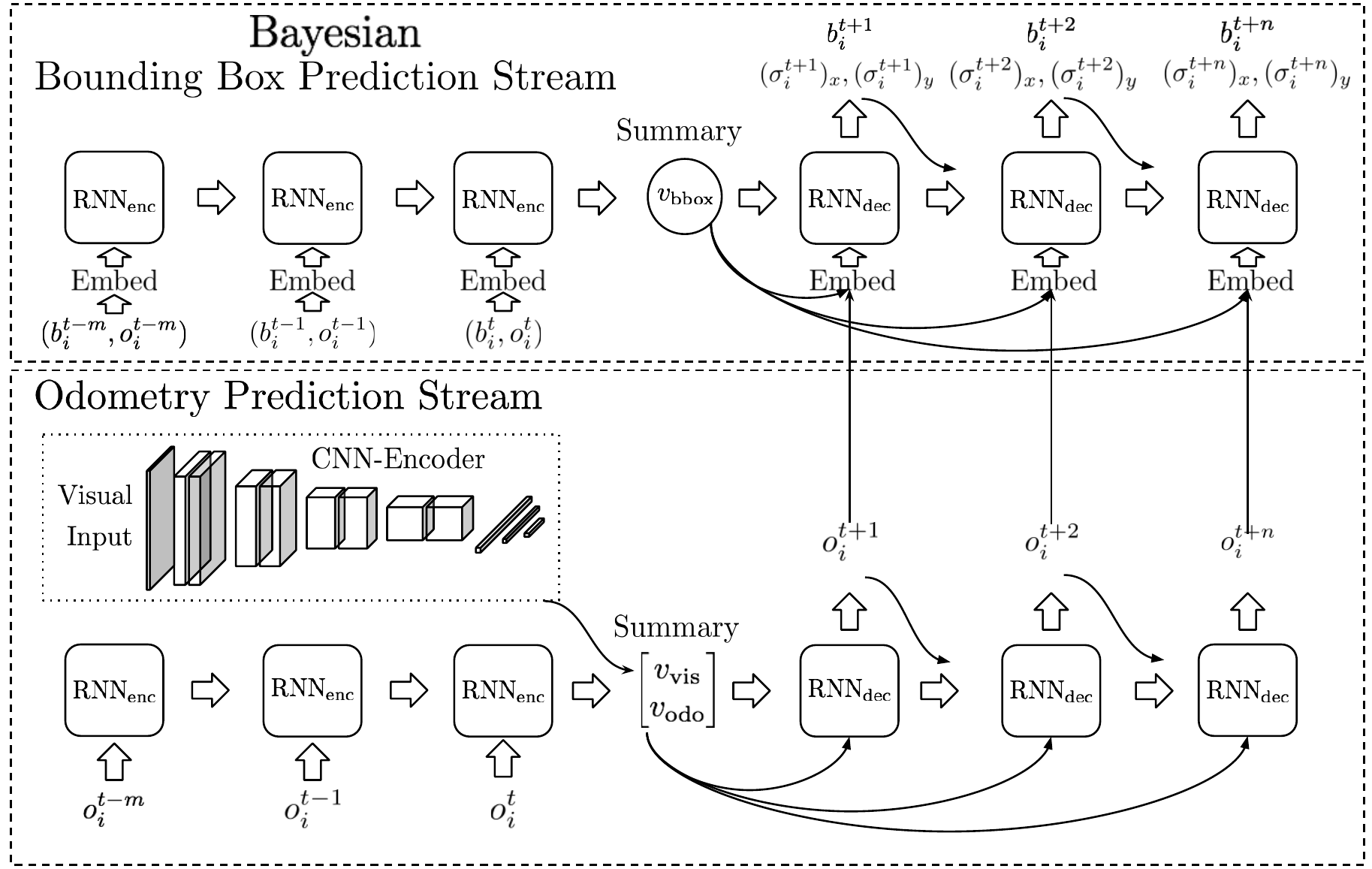}
    \end{subfigure}
    \caption{Two stream architecture for prediction of future pedestrian bounding boxes.}
    \vspace{-10pt}
    \label{fig:modelarch}
\end{figure*}

In order to anticipate motion of people in real-world traffic scenes from on-board cameras, we propose a novel approach that conditions the  prediction of motion (\autoref{sec:pedestrian_motion}) of people on predicted odometry (\autoref{sec:odometry}). Moreover, our approach models both \emph{aleatoric} and \emph{epistemic} uncertainty. Our model (see \autoref{fig:modelarch}) consists of two specialized streams for prediction of pedestrian motion and odometry. The odometry specialist stream predicts the most likely future vehicle odometry sequence. The bounding box specialist stream consists of a novel Bayesian RNN encoder-decoder architecture to predict odomerty conditioned  distributions over pedestrian trajectories and to capture epistemic and aleatoric uncertainty. Bayesian probability theory provides us with a theoretically grounded approach to dealing with both types of uncertainties (\autoref{sec:uncertainty}). %Our predictive distribution is richer and takes the form of a Gaussian mixture to capture epistemic uncertainty. 
%We base our model on RNN encoder-decoder architectures which are popular \cite{cho2014learning,sutskever2014sequence} for sequence to sequence learning tasks. 
 
We start by describing the bounding box prediction stream of our model and introduce our novel Bayesian RNN encoder-decoder which provides theoretically grounded uncertainty estimates.

\subsection{Prediction of Pedestrian Trajectories}\label{sec:pedestrian_motion}
A bounding box corresponding to the $i^{th}$ pedestrian observed on-board a vehicle at time step $t$ can be described by the top-left and bottom-right pixel coordinates: $b_{i}^{t} = \left\{(x_{tl},y_{tl}),(x_{br},y_{br})\right\}$. We want to predict the distribution of future bounding box sequences $\text{B}_{\text{f}}$ (where $| \text{B}_{\text{p}}| = m$) of the pedestrian. We condition our predictions on the past bounding box sequence $\text{B}_{\text{p}}$, the past odometry sequence $\text{O}_{\text{p}}$ and the corresponding future odometry sequence $\text{O}_{\text{f}}$ of the vehicle. The future odometry sequence $\text{O}_{\text{f}}$ is predicted conditioned on the past odometry sequence $\text{O}_{\text{p}}$ and on-board visual observation. Odometry sequences consists of the speed $s^{t}$ and steering angle $d^{t}$ of the vehicle, that is, $o^{t} = (s^{t},d^{t})$.
\begin{align*}
    p(\text{B}_{\text{f}} = [ b_{i}^{t+1}&, ... , b_{i}^{t+n}] | \text{B}_{\text{p}}, \text{O}_{\text{p}}, \text{O}_{\text{f}} ) \\
    \text{B}_{\text{p}} &= [ b_{i}^{t-m}, ... , b_{i}^{t} ],\;\\
    \text{O}_{\text{p}} &= [  o^{t-m}, ... ,o^{t} ], \; \\
    \text{O}_{\text{f}} &= [ o^{t+1}, ... ,o^{t+n}] 
\end{align*}

 The variance of the predictive distribution $p(\text{B}_{\text{f}} | \text{B}_{\text{p}}, \text{O}_{\text{p}}, \text{O}_{\text{f}})$ provides a measure of the associated uncertainty.

 We will describe a basic sequence to sequence RNN first and then extend it to predict distributions and provide uncertainty estimates. Our  sequence to sequence RNN (\autoref{fig:modelarch}) consists of two embedding layers, an encoder RNN and a decoder RNN. The input sequence consists of the concatenated past bounding box and odometry sequences $\text{B}_{\text{p}}, \text{O}_{\text{p}}$. The input embedding layer embeds the inputs sequence $x_{t}$ into the representation $\hat{x}_{t}$.  This embedded  sequence is read by the encoder RNN ($\operatorname{RNN}_{\text{enc}}$) which produces a summary vector $v_{bbox}$. This summary vector is concatenated with predicted odometry $\text{O}_{\text{f}}$ and this summary sequence is embedded using the second embedding layer. This embedded summary sequence $\hat{v}$ (containing information about past pedestrian motion, past and future vehicle odometry) is used by the decoder RNN ($\operatorname{RNN}_{\text{dec}}$) for prediction. 

In the following, we extend this model to predict distributions and estimate uncertainty.

\iffalse
We can obtain such a point estimate through a linear transformation of the hidden state of the decoder RNN.
\begin{align}\label{eq12}
    \begin{split}
        h_{\text{dec}}^{t+n} &= \text{RNN}_{\text{dec}}(h_{\text{dec}}^{t+n-1}, v_{bbox}) \\ 
        \hat{b}_{i}^{t+n} &= W_{bbox} * h_{\text{dec}}^{t+n} + bias_{bbox}.
    \end{split}
\end{align}
\fi

\subsection{Bayesian Modelling of Uncertainty}\label{sec:uncertainty}
We phrase our novel RNN encoder-decoder model in a Bayesian framework \cite{kendall2017uncertainties}. We capture epistemic (model) uncertainty by learning a distribution of models $p(f | X,Y)$ likely to have generated our data $\left\{ X, Y\right\}$. Here, models $f$ are RNN encoder-decoders with varying parameters. We infer the posterior distribution of RNN encoder-decoders $p(f | X,Y)$ , given the prior belief of the distribution of RNN encoder-decoders $p(f)$. The predictive probability over the future sequence $\text{B}_{\text{f}}$ given the past sequence  $\text{B}_{\text{p}}$ is obtained by marginalizing over the posterior distribution of RNN encoder-decoders,
\begin{align}\label{eq1}
\begin{split}
p(\text{B}_{\text{f}} | \text{B}_{\text{p}},\text{O}_{\text{p}}, \text{O}_{\text{f}}, &X, Y) = \\
&\int p(\text{B}_{\text{f}} | \text{B}_{\text{f}}, , \text{O}_{\text{p}}, \text{O}_{\text{f}}, f) p(f | X, Y) df.
\end{split}
\end{align}

However, the integral in (\ref{eq1}) is intractable. But, we can approximate it in two steps \cite{Gal2016Bayesian,gal2016theoretically,kendall2017uncertainties}. First, we assume that our RNN encoder-decoder models can be described by a finite set of variables $\omega$. Thus, we constrain the set of possible RNN encoder-decoders to ones that can be described with $\omega$. Now, (\ref{eq1}) can be equivalently written as, 
\begin{align}\label{eq2m1}
\begin{split}
p(\text{B}_{\text{f}} | \text{B}_{\text{p}}, \text{O}_{\text{p}}, \text{O}_{\text{f}},& X, Y) = \\ &\int p(\text{B}_{\text{f}} | \text{B}_{\text{p}}, \text{O}_{\text{p}}, \text{O}_{\text{f}}, \omega) p(\omega | X, Y) d\omega
\end{split}
\end{align}

Second, we assume an approximating variational distribution $q(\omega)$ which allows efficient sampling,
\begin{align}\label{eq2}
p(\text{B}_{\text{f}} | \text{B}_{\text{p}}, \text{O}_{\text{p}}, \text{O}_{\text{f}}) = \int p(\text{B}_{\text{f}} | \text{B}_{\text{p}}, \text{O}_{\text{p}}, \text{O}_{\text{f}}, \omega) q(\omega) d\omega
\end{align}

We choose the set of weight matrices $\left\{ W_{1},..,W_{L}\right\}\in\mathcal{W}$ of our RNN enocder-decoder as the set of variables $\omega$. Then we define an approximating Bernoulli variational distribution $q(\omega)$ over the columns $w_{k}^{c}$ of the weight matrices $W_{k} \in \mathcal{W}$,
\begin{align}\label{eq4}
    \begin{split}
        &q(W_{k}) = M_{k} \cdot \text{diag}(\lbrack z_{i,j}\rbrack^{C_{k}}_{j=1})\\
        z_{i,j} = &\;\text{Bernoulli}(p_{i}), i = 1, ..., L, j = 1, ..., K_{i - 1}.
    \end{split}
\end{align}
where, $M_{k}$ are the variational parameters. This distribution allows for efficient sampling during training and testing which we discuss in the following subsection.

For an accurate approximation, we minimize the KL divergence between $q(\omega)$ and the true posterior $p(\omega | X,Y)$ as the training step. It can be shown that, %(as in \cite{gal2016dropout,Gal2016Bayesian}),
\begin{align}\label{eq3}
    \begin{split}
        \text{KL}&(q(\omega) \mid\mid p(\omega | X,Y)) \propto \,\text{KL}(q(\omega)\mid\mid p(\omega)) \\
        - &\sum_{t} \int q(\omega) \log p(b_{t}^{t+n} | b_{t}^{t+n-1}, \text{B}_{\text{p}}, \text{O}_{\text{p}}, \text{O}_{\text{f}}, \omega) d\omega.
    \end{split}
\end{align}
The first part corresponds to the distance to the prior model distribution and the second to the data fit. During training and prediction, we use Monte-Carlo integration to approximate the integrals (\ref{eq2}) and (\ref{eq3}) (more details about (\ref{eq3}) in the Supplementary and the exact objective in \autoref{ssec:cl}).

Aleatoric uncertainty can be captured along with epistemic uncertainty, by assuming a distribution of observation noise and estimating the sufficient statistics of the distribution. Here, we assume it to be a 4-d Gaussian at each time-step, $\mathcal{N}(b_{i}^{t+n},\Sigma_{i}^{t+n})$, where, $\Sigma_{i}^{t+n}=\text{diag}\big((\sigma_{x}^{t+n})_{i},(\sigma_{y}^{t+n})_{i},(\sigma_{x}^{t+n})_{i},(\sigma_{y}^{t+n})_{i}\big)$ in $x$ and $y$ directions in pixel space at time-step $t+n$. The predictive distribution of models parametrized by $\omega$,  $p(\text{B}_{\text{f}} | \text{B}_{\text{p}}, , \text{O}_{\text{p}}, \text{O}_{\text{f}}, \omega)$ is Gaussian at every time-step.

Uncertainty is the variance of our predictive distribution (\ref{eq2}) and can be obtained through moment matching \cite{gal2016dropout,kendall2017uncertainties}. If we have $T$ samples of future pedestrian bounding box sequences $\hat{\text{B}}_{\text{f}}$, the total uncertainty at time-step $t$ is,  
\begin{align}\label{eq6}
\begin{split}
\frac{1}{T} \Big( \sum_{i=1}^{T} (\hat{b}^{t}_{i})^\intercal \hat{b}^{t}_{i} - \frac{1}{T} \big( \sum_{i=1}^{T} (\hat{b}^{t}_{i})^\intercal \big) \big( \sum_{i=1}^{T} \hat{b}^{t}_{i} \big) \Big) \\+ \frac{1}{T} \Big( \sum_{i=1}^{T} (\hat{\sigma}_{i}^{t})_{x} +\sum_{i=1}^{T} (\hat{\sigma}_{i}^{t})_{y} \Big).
\end{split}
\end{align}
The first part of the sum correspond to the epistemic uncertainty $u_{i}^{e}$ and the second part corresponds to the aleatoric uncertainty $u_{i}^{a}$. We average the uncertainty across time-steps to arrive at the complete uncertainty estimate.
\iffalse
The predictive distribution can be evaluated by measuring the probability assigned to the true $y^{*}$ (true $\text{B}_{\text{future}}$) using the negative log-likelihood ($\operatorname{\mathcal{L}}$) metric \cite{gal2016dropout},
\begin{align}
\operatorname{\mathcal{L}}(y^{*}) = -\log(\frac{1}{T}\sum_{i=1}^{T} p(y^{*} | x^{*}, \hat{\omega}_{i} )),\; \hat{\omega}_{i} \sim q(\omega).
\end{align}
\fi
Next, we describe how we sample from the Bernoulli distribution of RNN encoder-decoder weight matrices and the final sampling from the predictive distribution $p(\text{B}_{\text{f}} | \text{B}_{\text{p}}, \text{O}_{\text{p}}, \text{O}_{\text{f}} )$.

\subsection{Bayesian RNN Encoder-Decoder}
The RNN encoder-decoder model of \autoref{sec:pedestrian_motion} contains four weight matrices. In detail, the two embedding layers contains two weight matrices $W_\text{\it emi},W_\text{\it ems}$. The other two weight matrices belong to the encoder and decoder RNNs. We use an LSTM formulation as RNNs. Following \cite{graves2013speech} the weight matrices of an LSTM can be concatenated into a matrix $W$ and the LSTM can be formulated as in, 
\begin{align}\label{eq7}
    \begin{split}
        \begin{pmatrix} \underline{i} \\ \underline{f} \\ \underline{o} \\ \hat{\underline{c}} \end{pmatrix}
        = \begin{pmatrix} \text{sigm} \\ \text{sigm} \\ \text{sigm} \\ \text{tanh} \end{pmatrix}
        \begin{pmatrix} {\begin{pmatrix} \hat{x}_{t} \\ h_{t - 1}  \end{pmatrix}} \cdot W \end{pmatrix}\\
        c_{t} = \underline{f} \odot c_{t-1} + \underline{i} \odot \hat{\underline{c}}\;, \;\;\; h_{t} = \underline{o} \odot \text{tanh}(c_{t})
    \end{split}
\end{align}

%are fully-connected with non-linearities. Therefore, the linear transformations ($\hat{x}_{t} = x_{t} \cdot W_{emi}$) are repeated over all time-steps of the input sequence and summary sequence

where ${\underline{i}}$ is the input gate, ${\underline{f}}$ is the forget gate, ${\underline{o}}$ is the output gate, ${c_{t}}$ is the cell state, ${\hat{\underline{c}}}$ is the candidate cell state and ${h_{t}}$ is the hidden state. 

%The encoder LSTM with weight matrix $W_{enc}$ reads in the embedded input sequence by applying (\ref{eq7}) recurrently at each time step. The hidden state $h_{t}$ at the end of reading the input sequence is the summary vector $v_{bbox}$ (\autoref{fig:modelarch}). The decoder LSTM with weight matrix $W_{dec}$ reads the embedded summary $v_{bbox}$ as the input at every time-step and recurrently applies (\ref{eq7}).

We define the Bernoulli variational distribution $q(\omega)$ (as in (\ref{eq4})) over the union of all the weight matrices of our model,
\begin{align}\label{eq8}
\omega = \left\{ W_\text{\it emi}, W_\text{\it ems}, W_\text{\it enc}, W_\text{\it dec} \right\}.
\end{align}
where, $W_\text{\it enc},W_\text{\it dec}$ are the weight matrices of our RNN encoder and decoder.

Sampling from $q(W_\text{\it emi}),q(W_\text{\it ems})$ can be done efficiently by sampling random Bernoulli masks $z_{emi},z_{ems}$ and applying these masks after the linear transformations. In case of the input embedding,
\begin{align}\label{eq9}
\hat{x}_{t} = (x_{t} \cdot W_\text{\it emi}) \odot z_\text{\it emi}
\end{align}
Similarly, it was shown in \cite{gal2016theoretically} sampling weight matrices of a LSTM (here, $q(W_\text{\it enc}),q(W_\text{\it dec})$) can be efficiently performed by sampling random Bernoulli masks $z_{x},z_{h}$ and applying them at each time-step, while the LSTM encoder and decoder are unrolled,
\begin{align}\label{eq10}
\begin{pmatrix} \underline{i} \\ \underline{f} \\ \underline{o} \\ \hat{\underline{c}} \end{pmatrix}
= \begin{pmatrix} \text{sigm} \\ \text{sigm} \\ \text{sigm} \\ \text{tanh} \end{pmatrix}
\begin{pmatrix} {\begin{pmatrix} x_{t} \odot z_{x} \\ h_{t - 1} \odot z_{h}  \end{pmatrix}} \cdot W \end{pmatrix}
\end{align}

Sampling from our predictive distribution $p( \text{B}_{\text{f}} |  \text{B}_{\text{p}}, \text{O}_{\text{f}}, \text{O}_{\text{p}} )$ is done by first sampling weights matrices of our Bayesian RNN encoder-decoder. Then the parameters of the Gaussian observation noise distribution at each time-step is predicted. For this, we use the hidden state sequence $h_{\text{dec}}^{t}$ of the $\operatorname{RNN}_{\text{dec}}$ and an additional linear transformation,
\begin{align*}
    \begin{split}
        h_{\text{dec}}^{t+n} &= \text{RNN}_{\text{dec}}(h_{\text{dec}}^{t+n-1}, v_{bbox}; z_{x}, z_{h}) \\ 
        \hat{b}_{i}^{t+n},\;(\hat{\sigma_{i}}^{t+n})_{x},\;&(\hat{\sigma}_{i}^{t+n})_{y} = W_{bbox} * h_{\text{dec}}^{t+n} + bias_{bbox}.
    \end{split}
\end{align*}
We then draw a sample from the predicted Gaussian distribution. 

\iffalse
Thus, our predictive distribution factorizes over time-steps and the predictive distribution at time $t+n$ is conditioned upon the summary vector $v_{bbox}$ (thus $\text{B}_{\text{past}}$), the predicted odometry and the distribution at the previous time-step. \mario{you always talk about ``can'' in the previous paragraph. say what you do - otherwise it is perceived as optional}
\fi

Next, we describe the second stream of our two-stream model -- our model for long-term odometry prediction.

\subsection{Prediction of Odometry}\label{sec:odometry}
The odomtery prediction stream predicts a mean estimate of the future vehicle ego-motion. We use a similar RNN encoder-decoder architecture used for bounding box prediction, but without the embedding layers. We condition the predicted sequence $\text{O}_{\text{f}}$ on the past odometry sequence $\text{O}_{\text{p}}$ and last visual observation on-board the vehicle. The past odometry $\text{O}_{\text{p}}$ is input to an encoder RNN which produces a summary vector $v_{odo}$. The past odometry of the vehicle $\text{O}_{\text{p}}$ gives a strong cue about the future velocity especially in the short term ($\sim$100ms). We use the same LSTM formulation described previously as the RNN encoder; with the final hidden state $h^{t}$ as the summary. The last visual observation can help in the longer term prediction of odometry; e.g. visual cues about bends in the road, obstacles etc. Similar to \cite{xu2016end,bojarski2016end} we employ a convolutional neural network (CNN-encoder) to embed the visual information provided by the currently observed frame; a visual summary vector $v_{vis}$. Next we describe our CNN-encoder architecture.

\myparagraph{CNN-encoder.}
Our CNN-encoder should extract visual features to improve longer-term (multi-step versus single-step in \cite{xu2016end,bojarski2016end}) prediction. Therefore, we use a more complex CNN compared to \cite{bojarski2016end} and during training we learn the parameters from scratch, unlike \cite{xu2016end} which uses a pre-trained VGG network. Our CNN-encoder has 10 convolutional layers with \emph{ReLU} non-linearities. We use a fixed, small filter size of 3x3 pixels. We use max-pooling after every two layers. After max-pooling we double the number of convolutional filters; we use \{32,64,128,256,512\} convolutional filters. The convolutional layers are followed by three fully connected layers with 1024, 256 and 128 neurons and \emph{ReLU} non-linearities. The output of the last fully connected layer is the visual summary $v_{vis}$.

The odometry and visual summary vectors  are concatenated $v = \left\{ v_{odo}, v_{vis}\right\}$ and read by the RNN decoder ($\operatorname{RNN}_{\text{dec}}$). We use the same LSTM formulation described previously as the RNN-decoder. As before, the hidden state of the LSTM decoder is used for predicting the future odometry sequence through a linear transformation.
\begin{align*}
    h_{\text{dec}}^{t+n} &= \text{RNN}_{\text{dec}}(h_{\text{dec}}^{t+n-1}, \left\{v_{odo},v_{vis}\right\}) \\ 
    o_{i}^{t+n} &= W_{odo} * h_{\text{dec}}^{t+n} + bias_{odo}.
\end{align*}
We next describe our training and inference processes.

\subsection{Training and Inference}\label{ssec:cl} 
\myparagraph{Training.} The two streams are trained separately. As the odometry prediction stream predicts point estimates, it is trained first by minimizing the MSE over the training set. The Bayesian bounding-box prediction stream is trained by estimating (Monte-Carlo) and minimizing the KL divergence of its approximate weight distribution $q(\omega)$ (\ref{eq3}). More specifically, 
\begin{enumerate*}
    \item We sample a mini-batch of size $T$ of examples from the training set.
    \item For each example, weights $\left\{ W_\text{\it emi}, W_\text{\it ems}, W_\text{\it enc}, W_\text{\it dec} \right\}$ are sampled from $q(\omega)$ (\ref{eq8}), by sampling Bernoulli masks as in (\ref{eq9}) and (\ref{eq10}).
    \item For each example, the predicted means $\hat{\text{B}}_{\text{f}}$ and variances $\hat{\sigma}$ of the heteroscedastic models parameterized by $\omega$ are inferred.
    \item The KL divergence (\ref{eq3}) can be equivalently minimized by (similar to \cite{gal2016dropout,kendall2017uncertainties}) the following loss,
\end{enumerate*}
\begin{align*}
\frac{1}{4N} \sum_{i=1}^{N} \sum_{j=1}^{n}  \lVert \hat{b}_{i}^{t+j} - b_{i}^{t+j} \lVert_{2}^{2} (\hat{\Sigma}_{i}^{t+j})^{-2} + \lambda \sum_{\mathcal{W}} \lVert W_{k} \rVert_{2} + \log \hat{\sigma}_{i}^{2}
\end{align*}
where, $|\text{B}_{\text{f}}| = n$ and N pedestrians. The left part is the equivalent of the negative log likelihood term in (\ref{eq3}). The middle part is weight regularization parameterized by $\lambda$, equivalent to the KL term in (\ref{eq3}). The right part is additional regularization as in \cite{kendall2017uncertainties}, to ensure finite predicted variance. 

The ADAM optimizer \cite{kingma2014adam} is used during training. For training sequences longer than $| \text{B}_{\text{p}}| + | \text{B}_{\text{f}}|$ ($| \text{O}_{\text{p}} + \text{O}_{\text{f}}|$ respectively) we use a sliding window to convert to multiple sequences. Moreover, as the sequences in the training set are of varying lengths, 
%with longer sequences rarer than shorter sequences, 
we use a curriculum learning (CL) approach. We fix the length of the conditioning sequence $| \text{B}_{\text{p}}|,| \text{O}_{\text{p}}|$ and train for increasing longer time horizons $| \text{B}_{\text{f}}|,| \text{O}_{\text{f}}|$ (initializing the model parameters with those for shorter horizons). This allows us to train on a larger part of the Cityscapes training set (also on sequences shorter than $| \text{B}_{\text{p}}| + | \text{B}_{\text{f}}|$ of the final model) and leads to faster convergence.

\myparagraph{Inference.} Given $\text{B}_{\text{p}}$ and $\text{O}_{\text{p}}$ (and the visual observation), the odometry prediction stream is first used to predict $\text{O}_{\text{f}}$. We sample from the predictive distribution (\ref{eq2}) by, 
\begin{enumerate*}
\item Sampling $T$ samples of the weight matrices $\left\{ W_\text{\it emi}, W_\text{\it ems}, W_\text{\it enc}, W_\text{\it dec} \right\}$ of the Bayesian bounding box prediction stream from the (learned) approximate distribution $q(\omega)$, by sampling Bernoulli masks as in (\ref{eq9}) and (\ref{eq10}),
\item The $\text{RNN}_{\text{dec}}$ is unrolled to obtain a sample $\left\{ \hat{\text{B}}_{\text{f}}, \hat{\sigma}_{x}, \hat{\sigma}_{y} \right\}$ from each of the $T$ predicted Gaussian distributions.
\end{enumerate*}
The associated uncertainty is obtained using the $T$ samples (\ref{eq6}).

\section{Experiments}\label{sec:exp}
We evaluate our model on real-world on-board street scene data and show predictions over a 1 second time horizon along with the associated uncertainty.

\begin{table*}[t]
\centering
\setlength\tabcolsep{4pt}
\begin{minipage}{0.67\textwidth}
\centering

\begin{tabular}{cccccccc}
\toprule
&& \multicolumn{3}{c}{MSE} & \multicolumn{3}{c}{$\operatorname{\mathcal{L}}$} \\
\cmidrule{3-8}
&& \multicolumn{3}{c}{$| \text{B}_{\text{p}}|$} & \multicolumn{3}{c}{$| \text{B}_{\text{p}}|$} \\
 Method & Odometry & 4 & 6 & 8 & 4 & 6 & 8 \\
\midrule
Kalman Filter & None & 1938 & 1289 & 1098 & x & x & x \\ 
LSTM & None & 692 & 663 & 650 & 8.11 & 7.99 & 7.77 \\
LSTM-Aleatoric & None & 772 & 758 & 750 & 5.92 & 5.81 & 5.54 \\
LSTM-Bayesian & None & \textbf{647} & \textbf{624} & \textbf{618} & \textbf{4.31} & \textbf{4.26} & \textbf{4.13} \\
\midrule
LSTM-Bayesian & Ground-truth & 374 & 358 & 343 & 3.94 & 3.93 & 3.88\\ 
\bottomrule
\end{tabular}
\caption{Bounding box prediction error with varying $| \text{B}_{p}|$.}
\label{tab:stream1eval}
\end{minipage}%
\hfill
\begin{minipage}{0.32\textwidth}
\centering
\begin{tabular}{ccc}
\toprule
Method &  MSE & $\operatorname{\mathcal{L}}$\\
\midrule
Social LSTM \cite{alahi2016social} & 1514 & 5.63 \\
LSTM-Bayesian & 695 & 3.97 \\
LSTM-Bayesian (centers) & 648 & x \\
\bottomrule
\end{tabular}
\caption{Bounding box center prediction error.}
\label{tab:bboxcenter}
\end{minipage}
\end{table*}

\myparagraph{Dataset and Evaluation Metric.}
We evaluate on the Cityscapes dataset \cite{Cordts2016Cityscapes} which contains 2975 training, 500 validation and 1525 test video sequences of length 1.8 seconds (30 frames). The video resolution is 2048$\times$1024 pixels. The sequences were recorded on-board a vehicle in inner cities. Each sequence has associated odometry information. Pedestrian tracks were automatically extracted using the tracking by detection method of \cite{siyu}. Detections were obtained using the Faster R-CNN based method of \cite{zhang2017citypersons} (statistics in the Supplementary). This mimics real world autonomous/assisted driving systems where detections/tracks are obtained with a state-of-the-art detector/tracker and we have to deal with noise introduced by the detector and on rare occasions detector false positives and tracker failures. We use as evaluation metric MSE in pixels (of the mean of the predictive distribution) and the negative log-likelihood $\operatorname{\mathcal{L}}$. The $\operatorname{\mathcal{L}}$ metric measures the probability assigned to the true sequence by our predictive distribution. We report these metrics averaged across all time-steps and plots per time-step. We use a dropout rate of 0.35, $\lambda = 10^{-4}$ (tuned on validation set) and use 50 Monte-Carlo samples across all Bayesian models. 

\myparagraph{Evaluation of Bounding Box Prediction.} We independently evaluate the first Bayesian LSTM stream of our two stream model, without conditioning it on predicted odometry. We predict 15 time-steps into the future and report the results in \autoref{tab:stream1eval}. We compare its performance with,
\begin{enumerate*}
    \item A linear Kalman filter baseline.
    \item A LSTM encoder-decoder model which does not model uncertainty (LSTM).
    \item A LSTM encoder-decoder which models only aleatoric uncertainty (LSTM-Aleatoric).
\end{enumerate*}
 Finally, as an Oracle case, we compare against a Bayesian version in which the LSTM encoder can see the past odometry and the LSTM decoder can see the true future odometry at every time-step. We also vary the length of the conditioning sequence $| \text{B}_{\text{p}}|$ (training/test sets constant across varying $| \text{B}_{\text{p}}|$). In \autoref{tab:stream1eval}, we see that the homoscedastic LSTM model (2nd row) outperforms the linear Kalman filter (1st row).
 This shows that many bounding box sequences have a complex motion and therefore cannot be modelled by a Kalman filter. We see that the LSTM-Aleatoric (3rd row) outperforms the basic LSTM (2nd row) with respect to the $\operatorname{\mathcal{L}}$ metric. This means that the LSTM-Aleatoric learns to capture uncertainty and assigns higher probability to the true bounding box sequence. However, as epistemic uncertainty is not modelled, aleatoric uncertainty tries to compensate (as in \cite{kendall2017uncertainties}) and this leads to poorer MSE. Finally, our Bayesian LSTM (4th row) outperforms all other methods. This can be attributed to two factors,  
 \begin{enumerate*}
    \item The richer Gaussian mixture model fitted by the Bayesian LSTM can both capture aleatoric and epistemic uncertainty and fits the data distribution better (evidenced by $\operatorname{\mathcal{L}}$ metric).
    \item Additional introduced regularization.
 \end{enumerate*}
 Furthermore, we see that increasing the length of the conditioning sequence improves model performance. However, the performance gain saturates at $|\text{B}_{p}| = 8$. Henceforth, we will report results using $|\text{B}_{\text{p}}| = \left\{4,8\right\}$ in the following. Finally, the odometry oracle case outperforms our Bayesian LSTM by a large margin. This shows that knowledge of vehicle odometry is crucial for good performance. 

\myparagraph{Comparison with Social LSTM \cite{alahi2016social}.} We compare our Bayesian LSTM model with the vanilla LSTM \footnote{The version with social pooling did not converge on our dataset.} model of \cite{alahi2016social} (with 128 neurons) that predicts trajectories independently in \autoref{tab:bboxcenter}. Both models are trained to predict sequences of bounding box centers (length 15, given 8). Our Bayesian LSTM model performs better as it is more robust to mistakes during recursive prediction. The model of \cite{alahi2016social} observes true past pedestrian coordinates during training. However, during prediction it observes its own predictions causing errors to be propagated though multiple steps of prediction. Furthermore, we compare both methods to the centers obtained from the predictions of our Bayesian LSTM (second row of \autoref{tab:stream1eval}). The results show that we can improve upon bounding box center prediction by predicting bounding boxes.

\begin{table}[!t]
\centering
\resizebox{\linewidth}{!}{
\begin{tabular}{cccc}
\toprule
Method & Visual & Speed (m/sec)  & Angle (degrees)\\
\midrule
Constant & None & 1.62 & 26.85\\
Kalman Filter & None & 0.053 & 2.44\\
LSTM & None & 0.056 & 0.94\\
LSTM & RGB & 0.048 & 0.88\\
\bottomrule
\end{tabular}
}
\caption{Odometry prediction error (MSE), $| \text{O}_{\text{p}}| = \left\{8\right\}$.}
\label{fig:stream2eval}
\end{table}

\begin{table*}
\begin{minipage}[b]{0.65\linewidth}
\centering
\begin{tabular}{ccccccc}
\toprule
&&& \multicolumn{2}{c}{MSE} & \multicolumn{2}{c}{$\operatorname{\mathcal{L}}$} \\
\cmidrule{4-7}
&&& \multicolumn{2}{c}{$| \text{B}_{\text{p}}|$} & \multicolumn{2}{c}{$| \text{B}_{\text{p}}|$} \\
 Method & Streams  & Visual & 4 & 8 & 4 & 8 \\
\midrule
Kalman Filter & x & None & 1938 & 1098 & x & x\\
LSTM-Bayesian & One & None & 572 & 546 & 4.03 & 3.97\\
LSTM-Bayesian & Two & RGB & \textbf{532} & \textbf{505} & \textbf{3.99} & \textbf{3.92}\\
\bottomrule
\end{tabular}
\caption{Evaluation of our Bayesian two stream model (\autoref{fig:modelarch}).}
\label{tab:bothstreameval}
\end{minipage}
\hspace{0.5cm}
\begin{minipage}[b]{0.32\linewidth}
\centering
\includegraphics[width=0.8\textwidth, keepaspectratio]{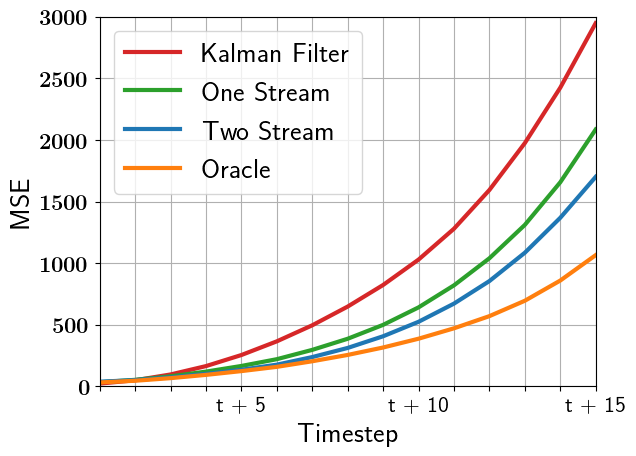}
\caption{MSE per time-step of models in \autoref{tab:stream1eval} row 1, 4, 5 and \autoref{tab:bothstreameval} row 3.}
\label{tab:perstepmse}
\end{minipage}
\end{table*}

\begin{figure*}[h]

\resizebox{\textwidth}{!}{
\begin{tabular}{cccc}

    \includegraphics[width=0.245\textwidth, height=0.15\textheight]{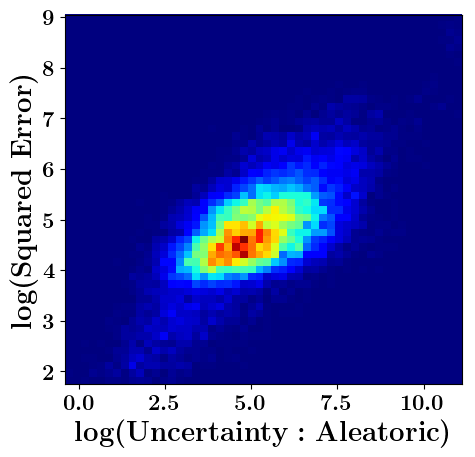} &
    \includegraphics[width=0.245\textwidth, height=0.15\textheight]{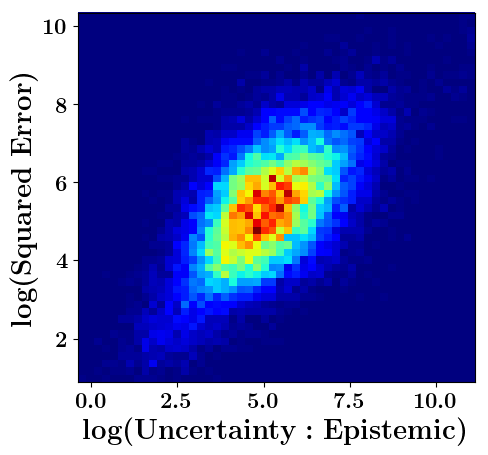} &
    \includegraphics[width=0.245\textwidth, height=0.15\textheight]{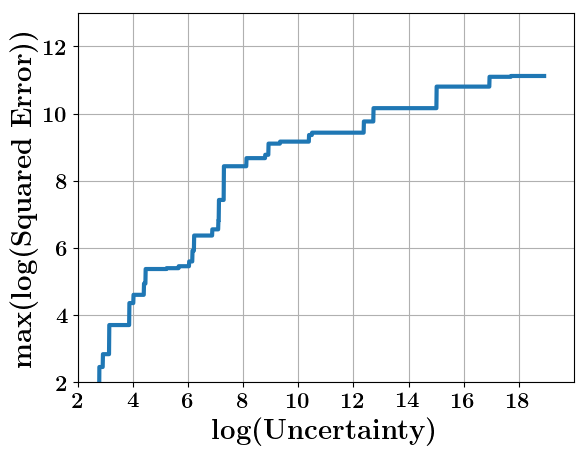} &
    \includegraphics[width=0.245\textwidth, height=0.15\textheight]{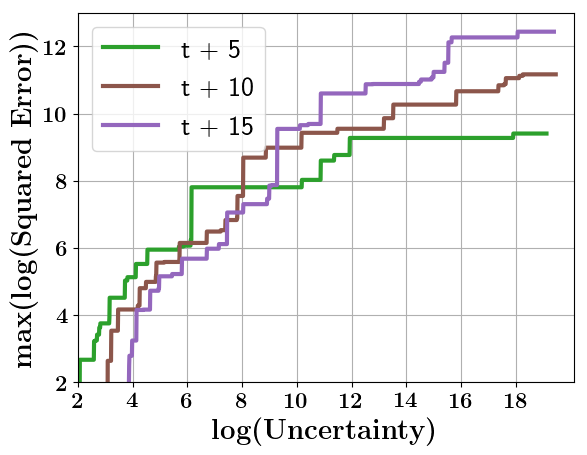}\\

\end{tabular}
}
\caption{Quality of our uncertainty metric: plots 1 and 2 - uncertainty versus squared error, plots 3 and 4 - uncertainty versus \emph{maximum} observed squared error.}\label{fig:uncertinitymetric}
\end{figure*}

\myparagraph{Evaluation of Odometry Prediction.} We train our odometry prediction LSTM encoder-decoder on the visual and odometry data of the Cityscapes training set. As many sequences have close to zero steering angle, we augment the training set to improve prediction performance. We reflect the steering angle and flip last observed image left to right of sequences with non-zero average steering angle. This increases the training data with non-zero steering angles by a factor of two.  We use MSE between the predicted future vehicle velocity and steering angles as evaluation metric. The velocity is in meters per second and angle in degrees. We include as baselines:
\begin{enumerate*}
    \item A constant steering predictor that predicts the last observed odometry.
    \item A linear Kalman filter.
    \item Our LSTM encoder-decoder without visual observation ($v = \left\{ v_{odo}\right\}$).
\end{enumerate*}
The third baseline is an ablation study. We observe no significant performance difference between $| \text{O}_{\text{p}}| = \left\{4\right\}$ and $| \text{O}_{\text{p}}| = \left\{8\right\}$. We evaluate 15 time-steps into the future and report the results in \autoref{fig:stream2eval}. We observe that the constant angle predictor performs significantly worse compared to the other baselines. This shows that the Cityscapes test set includes a significant number of non-trivial sequences with complex vehicle trajectories. We observe that the Kalman filter is able to quite accurately predict the vehicle speed. This is because in most vehicles are travelling with constant speed or accelerating/decelerating smoothly. However, the performance of the linear Kalman filter is worse compared to the LSTM models with respect to steering angle. This means that many sequences have non-linear vehicle trajectories. The superior performance of our model compared to the RNN baseline without visual observations, especially in the long-term shows that our CNN encoder extracts information useful for long-term prediction. We also show visual examples in the Supplementary.

\begin{figure*}[!t]

\resizebox{\textwidth}{!}{
\begin{tabular}{cccc}
\toprule

Last Observation: $t$ & Prediction: $t$ + 5  & Prediction: $t$ + 10  & Prediction: $t$ + 15 \\

\midrule

    \includegraphics[width=0.18\textwidth, height=0.072\textheight]{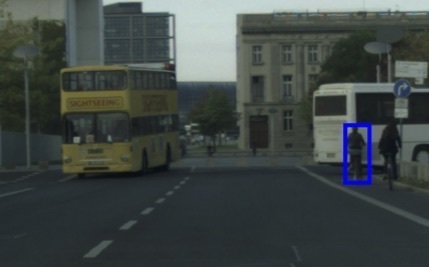} &
    \includegraphics[width=0.18\textwidth, height=0.072\textheight]{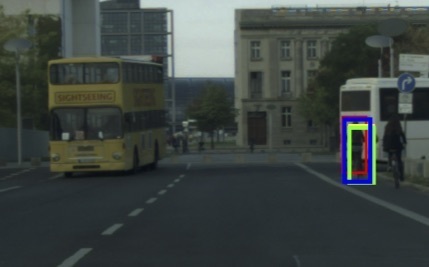} &
    \includegraphics[width=0.18\textwidth, height=0.072\textheight]{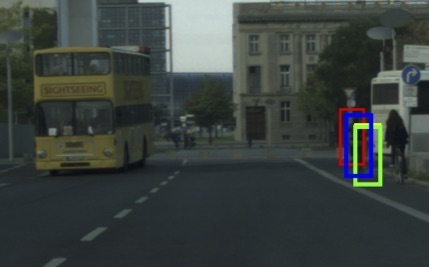} &
    \includegraphics[width=0.18\textwidth, height=0.072\textheight]{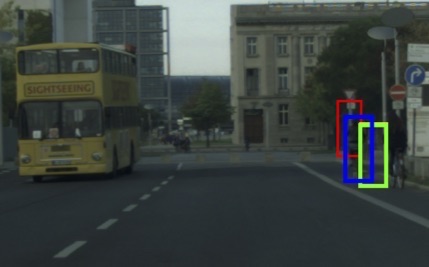}\\
    \addlinespace[-0.5ex]
    \includegraphics[width=0.18\textwidth, height=0.072\textheight]{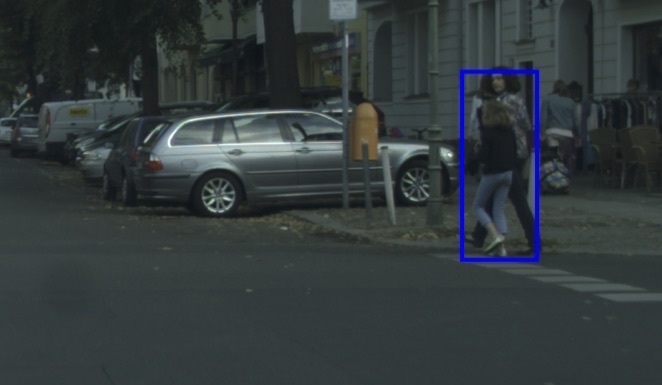}&
    \includegraphics[width=0.18\textwidth, height=0.072\textheight]{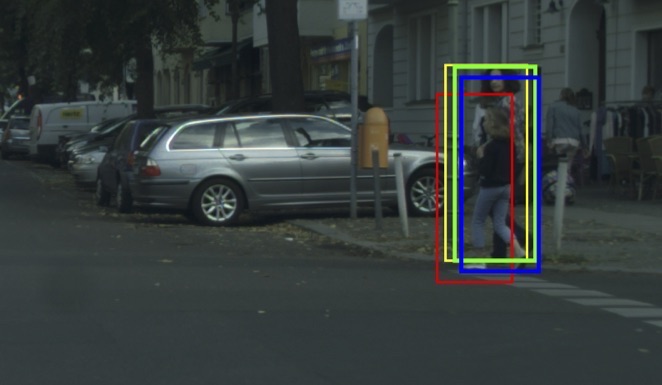}&
    \includegraphics[width=0.18\textwidth, height=0.072\textheight]{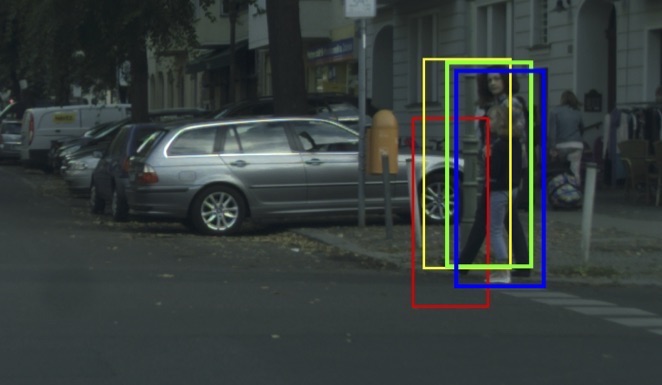}&
    \includegraphics[width=0.18\textwidth, height=0.072\textheight]{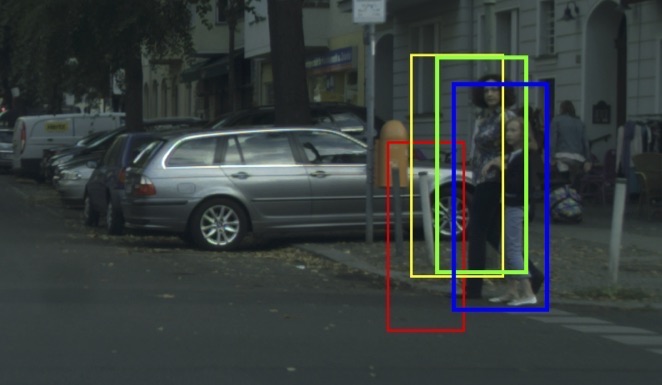} \\
    \addlinespace[-0.5ex]
    \includegraphics[width=0.18\textwidth, height=0.072\textheight]{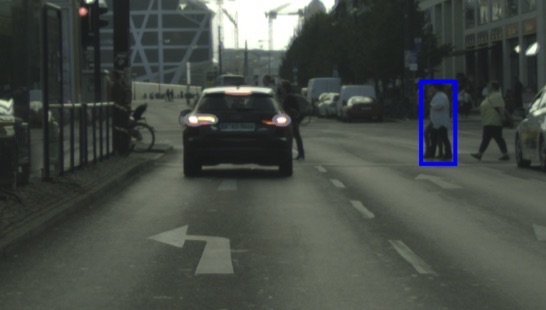}&
    \includegraphics[width=0.18\textwidth, height=0.072\textheight]{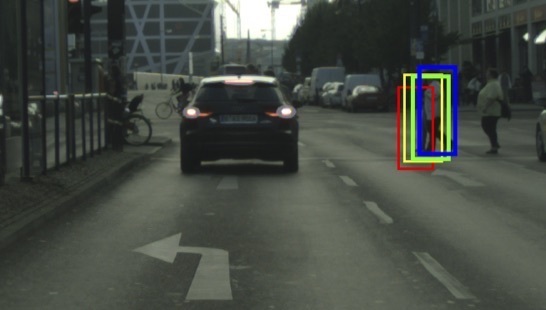}&
    \includegraphics[width=0.18\textwidth, height=0.072\textheight]{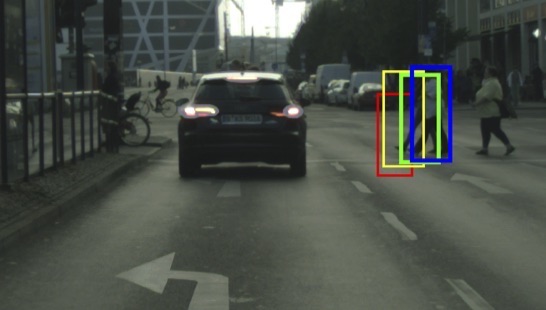}&
    \includegraphics[width=0.18\textwidth, height=0.072\textheight]{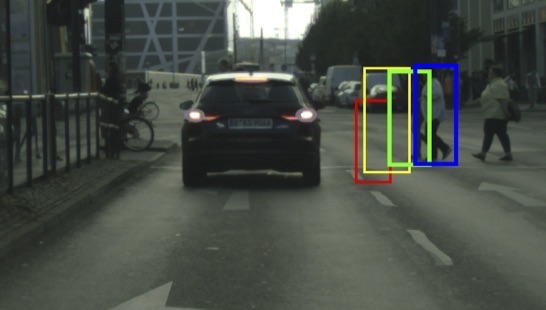} \\

\hline
\addlinespace[-2.4ex]
    \\

    \includegraphics[width=0.18\textwidth, height=0.072\textheight]{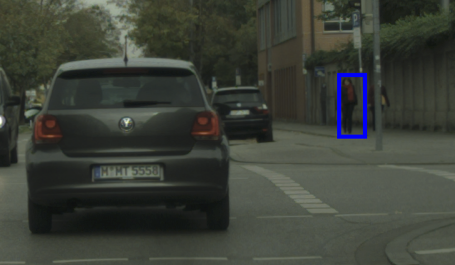} &
    \includegraphics[width=0.18\textwidth, height=0.072\textheight]{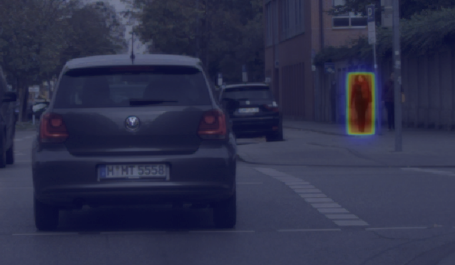} &
    \includegraphics[width=0.18\textwidth, height=0.072\textheight]{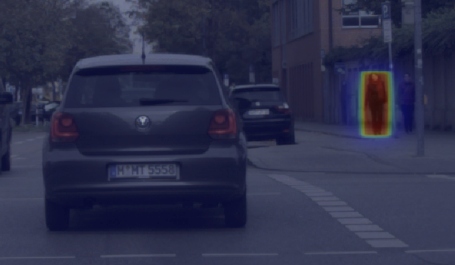} &
    \includegraphics[width=0.18\textwidth, height=0.072\textheight]{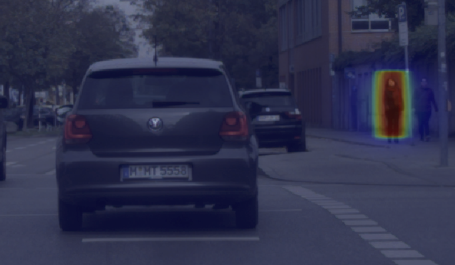}\\
    \addlinespace[-0.5ex]
    \includegraphics[width=0.18\textwidth, height=0.072\textheight]{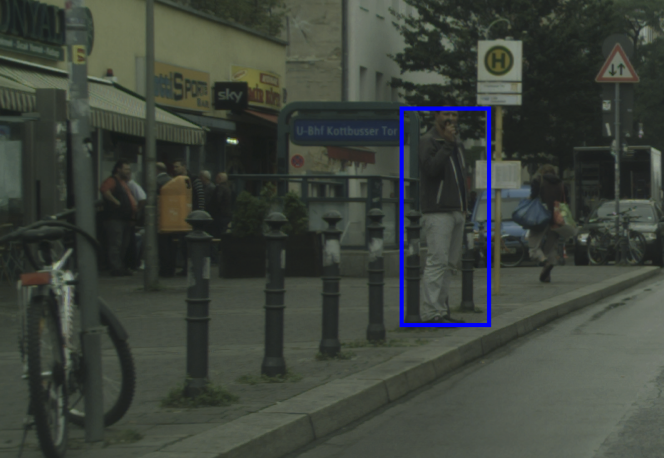}&
    \includegraphics[width=0.18\textwidth, height=0.072\textheight]{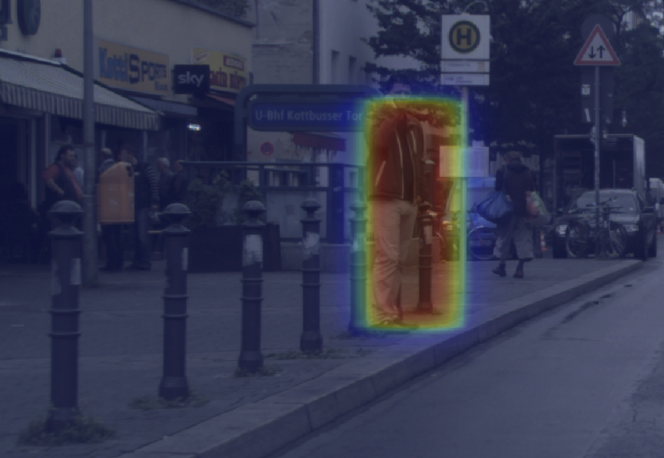}&
    \includegraphics[width=0.18\textwidth, height=0.072\textheight]{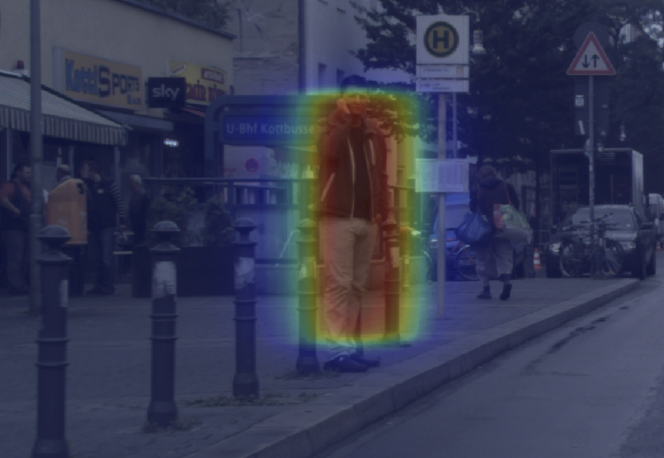}&
    \includegraphics[width=0.18\textwidth, height=0.072\textheight]{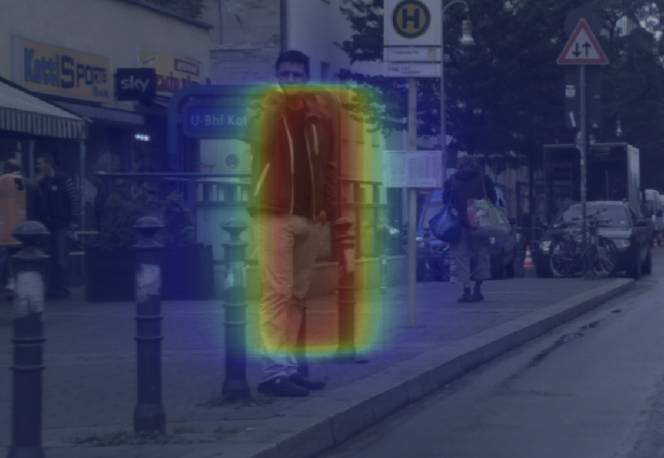} \\
    \addlinespace[-0.5ex]
    \includegraphics[width=0.18\textwidth, height=0.072\textheight]{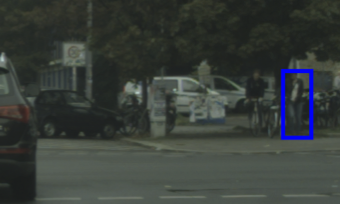}&
    \includegraphics[width=0.18\textwidth, height=0.072\textheight]{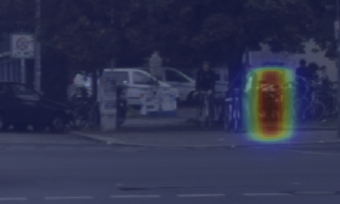}&
    \includegraphics[width=0.18\textwidth, height=0.072\textheight]{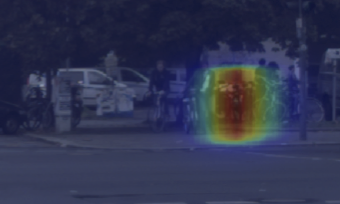}&
    \includegraphics[width=0.18\textwidth, height=0.072\textheight]{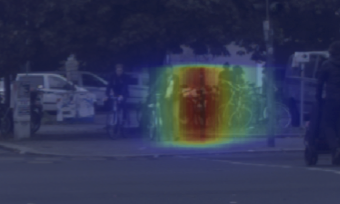} \\

\bottomrule
\end{tabular}
}
\caption{\textbf{Rows 1-3}: Point estimates. \textcolor{black}{Blue}: Ground-truth, \textcolor{black}{Red}: Kalman Filter (\autoref{tab:stream1eval} row 1), \textcolor{black}{Yellow}: One-stream model (\autoref{tab:stream1eval} row 4), \textcolor{black}{Green}: Two-stream model (mean of predictive distribution, \autoref{tab:bothstreameval} row 3). \textbf{Rows 4-6:} Predictive distributions of our two-stream model as heat maps. (Link to video results in the Appendix). }\label{fig:uncertianityq}
\end{figure*}
%\emph{Row 1}: Case with linear vehicle ego-motion, all methods perform reasonably well. \emph{Rows 2 and 3}: cases with non-linear vehicle ego-motion, our two-stream model outperforms other methods.
%

\myparagraph{Evaluation of our Two-Stream model.}  We perform an ablation study of our two-stream model (\autoref{fig:modelarch}) and compare with a single-stream Bayesian LSTM encoder-decoder model where the encoder observes the concatenated past bounding box and velocity sequence $\left\{ \text{B}_{\text{p}}, \text{O}_{\text{p}}\right\}$ and the decoder predicts the future bounding box sequence $\text{B}_{\text{f}}$. This model does not see predicted future odometry. We evaluate the models and report the results in \autoref{tab:bothstreameval} and plot the MSE per time-step \autoref{tab:perstepmse}. The results show that jointly predicting odometry with pedestrian bounding boxes (3rd row) significantly improves performance (2nd row). The predicted odometry helps our two-stream model recover a significant fraction of the performance of the Oracle case in \autoref{tab:stream1eval} row 5. The limiting factor here is that the odometry is sometimes highly uncertain e.g. at T-intersections, which leads to higher mean error. Apart from cases with uncertain odometry, the residual error of our two-stream (and the Oracle case) on a large part is due to the noise of the pedestrian detector and tracker failures. We show qualitative examples in \autoref{fig:uncertianityq}. Row 1 shows point estimates under linear vehicle ego-motion and Rows 2, 3 non-linear vehicle ego-motion. Our two-stream model (mean of predictive distribution) outperforms other methods in the second case. Rows 4-5 shows the predictive distributions of the two-stream model under linear vehicle and pedestrian motion. The distribution is symmetric and has high aleatoric uncertainty which captures detection noise and possible pedestrian motion. Row 6 shows a case of a skewed distribution with high epistemic uncertainty which captures uncertainty in vehicle motion.

\myparagraph{Quality of our Uncertainty Metric.} We evaluate our uncertainty metric in \autoref{fig:uncertinitymetric}. The first two plots show the aleatoric and epistemic uncertainty to the squared error of the mean of the predictive distribution of our two-stream model. We use $\log$-$\log$ plots for better visualization as most sequences have low error (note, $\log(530) \approx 6.22$ the MSE of our two stream model, \autoref{tab:bothstreameval}). We see that the epistemic and aleatoric uncertainties are correlates well with the squared error. This means that for sequences where the mean of our predictive distribution is far from the true future sequence, our predictive distribution has a high variance (and vice versa). Therefore, for sequences with multiple likely futures, where the mean estimate would have high error, our model learns to predict diverse futures. In the third plot of \autoref{fig:uncertinitymetric}, we plot the \emph{maximum} $\log$ squared error (of the mean of the predictive distribution) observed at a certain predicted uncertainty level (sum of aleatoric and epistemic) in the test test. In the fourth plot, we plot the uncertainty with the maximum observed squared error at time-steps $t + \left\{5,10,15\right\}$. In both cases, uncertainty and observed maximum error is well correlated. This shows that, \emph{the predicted uncertainty upper bounds the error of the mean of the predictive distribution}. Therefore, the predicted uncertainty helps us express trust in predictions and has the potential to serve as a basis for better decision making. 

\section{Conclusion}
We highlight the importance of anticipation for practical and safe driving  in inner cities. We contribute to this important research direction the first model for long term prediction of pedestrians from on-board observations. We show  predictions over a time horizon of 1 second. Predictions of our model are enriched by theoretically grounded uncertainty estimates. Key to our success is a Bayesian approach and long term prediction of odometry. We evaluate and compare several different architecture choices and arrive at a novel two-stream Bayesian LSTM encoder-decoder.

%------------------------------------------------------------------------

{\small
\bibliographystyle{ieee}
\bibliography{egbib}
}
\newpage

\begin{appendix}
\addappheadtotoc

\title{
   \begin{center}
      \Large\textbf{Appendix}\\
      %\large\textit{A. Thor}
   \end{center}
}

\maketitle

\section{Additional Details of Training Objective}
As derived in \cite{gal2016dropout,Gal2016Bayesian}, in Bayesian Regression, the KL divergence between a approximate variational posterior $q(\omega)$ and the true posterior $p(\omega | X, Y)$ distribution of models likely to have generated our data is given by,
\begin{align}\tag{A1}\label{eqA1}
    \begin{split}
        \text{KL}&(q(\omega) \mid\mid p(\omega | X,Y)) \propto \,\text{KL}(q(\omega)\mid\mid p(\omega)) \\
        - & \int q(\omega) \log p(Y | X, \omega) d\omega.
    \end{split}
\end{align}
In our case, as we train our model to predict future bounding box sequences given the past bounding box sequence, past and future vehiche odometry, we have   $X=\left\{\text{B}_{\text{p}},\text{O}_{\text{f}},\text{O}_{\text{p}}\right\}$ and $Y=\left\{\text{B}_{\text{f}}\right\}$. Therefore, the KL divergence is given by,
\begin{align}\tag{A2}\label{eqA2}
    \begin{split}
        \text{KL}&(q(\omega) \mid\mid p(\omega | X,Y)) \propto \,\text{KL}(q(\omega)\mid\mid p(\omega)) \\
        - & \int q(\omega) \log p(\text{B}_{\text{f}} | \text{B}_{\text{p}},\text{O}_{\text{f}},\text{O}_{\text{p}}, \omega) d\omega.
    \end{split}
\end{align}
As the bounding box at time $t+n$ in $\text{B}_{\text{f}}$ is predicted conditioned on the bounding box at time $t+n-1$ and the past bounding box sequence, past and future vehiche odometry, by our Bayesian RNN Encoder-Decoder, the KL divergence is given by,  
\begin{align}\tag{A3}\label{eqA3}
    \begin{split}
        \text{KL}&(q(\omega) \mid\mid p(\omega | X,Y)) \propto \,\text{KL}(q(\omega)\mid\mid p(\omega)) \\
        - &\sum_{t} \int q(\omega) \log p(b_{t}^{t+n} | b_{t}^{t+n-1}, \text{B}_{\text{p}}, \text{O}_{\text{p}}, \text{O}_{\text{f}}, \omega) d\omega.
    \end{split}
\end{align}
During training (as mentioned in subsection 3.5 of the main paper), we use Monte-Carlo integration to estimate the integral in (\ref{eqA3}) (using $N$ samples),
\begin{align}\tag{A4}\label{eqA4}
    \begin{split}
        \text{KL}&(q(\omega) \mid\mid p(\omega | X,Y)) \propto \,\text{KL}(q(\omega)\mid\mid p(\omega)) \\
        - & \frac{1}{N} \sum_{t} \sum_{i=0}^{N} \log p(b_{t}^{t+n} | b_{t}^{t+n-1}, \text{B}_{\text{p}}, \text{O}_{\text{p}}, \text{O}_{\text{f}}, \hat{\omega}_{i}),\\ &\,\, \hat{\omega}_{i} \sim q(\omega).
    \end{split}
\end{align}
The probability term $p(b_{t}^{t+n} | b_{t}^{t+n-1}, \text{B}_{\text{p}}, \text{O}_{\text{p}}, \text{O}_{\text{f}}, \hat{\omega}_{i})$ takes the form $e^{-\lVert \hat{b}_{i}^{t+j} - b_{i}^{t+j} \lVert_{2}^{2} \, (\hat{\Sigma}_{i}^{t+j})^{-2}}$. Therefore, replacing the $\log$ probability term with the exponential squared error term and introducing additional regularization as mentioned in subsection 3.5 of the main paper leads to the training objective used,
\begin{align*}
    \frac{1}{4N} \sum_{i=1}^{N} \sum_{j=1}^{n}  \lVert \hat{b}_{i}^{t+j} - b_{i}^{t+j} \lVert_{2}^{2} (\hat{\Sigma}_{i}^{t+j})^{-2} + \lambda \sum_{\mathcal{W}} \lVert W_{k} \rVert_{2} + \log \hat{\sigma}_{i}^{2}
\end{align*}

\section{Additional Details of Two Stream Model}
Here, we include details of each layer of our Two Stream Model. We refer to fully connected layers as Dense and Size refers to the number of neurons in the layer.

\myparagraph{Bayesian Bounding Box Prediction Stream.} We provide the details of the Bayesian Bounding Box prediction stream in \autoref{tab:bbox_details1}. 

\begin{table}[h]
\centering
\resizebox{\linewidth}{!}{
\begin{tabular}{cccccc}
\toprule
Layer & Type & Size & Activation & Input & Output \\
\midrule
$\text{In}_1$ & Input & & & $\text{B}_{past}$ & $\text{EMB}_{1}$ \\
$\text{In}_2$ & Input & & & $\text{O}_{past}$ & $\text{EMB}_{1}$ \\
\midrule
$\text{EMB}_{1}$ & Dense & 64 & \emph{ReLU} & $\left\{\text{In}_1,\text{In}_2\right\}$ & $\text{LSTM}_{enc1}$ \\
$\text{LSTM}_{enc1}$ & LSTM & 128 & \emph{tanh} & $\text{EMB}_{1}$ & $\text{EMB}_{2}$ \\
\midrule
$\text{EMB}_{2}$ & Dense & 64 & \emph{ReLU} & $\left\{\text{LSTM}_{enc1},\hat{\text{O}}_{\text{f}}\right\}$ & $\text{LSTM}_{dec1}$ \\
$\text{LSTM}_{dec1}$ & LSTM & 128 & \emph{tanh} & $\text{EMB}_{2}$ & $\text{Out}_{1}$ \\
$\text{Out}_{1}$ & Dense & 4 & & $\text{LSTM}_{dec}$ & $\hat{\text{B}}_{\text{f}}$ \\
\bottomrule
\end{tabular}
}
\caption{Details of the Bounding Box Prediction Stream. Note that, the weights of all the layers are sampled from the approximate posterior $q(\omega)$.}
\label{tab:bbox_details1}
\end{table}

\myparagraph{Odometry Prediction Stream.} We provide the details of the odometry prediction stream in \autoref{tab:odo_details1}. We then provide details of the CNN encoder.

\begin{table}[h]
\centering
\resizebox{\linewidth}{!}{
\begin{tabular}{cccccc}
\toprule
Layer & Type & Size & Activation & Input & Output \\
\midrule
$\text{In}_3$ & Input & & & $\text{O}_{past}$ & $\text{LSTM}_{enc2}$ \\
\midrule
$\text{LSTM}_{enc2}$ & LSTM & 128 & \emph{tanh} & $\text{In}_{3}$ & $\text{LSTM}_{dec2}$ \\
\midrule
$\text{LSTM}_{dec2}$ & LSTM & 128 & \emph{tanh} & $\left\{\text{LSTM}_{enc1},\text{FC}_{3}\right\}$ & $\text{Out}_{1}$ \\
$\text{Out}_{2}$ & Dense & 2 & & $\text{LSTM}_{dec2}$ & $\hat{\text{O}}_{\text{f}}$ \\
\bottomrule
\end{tabular}
}
\caption{Details of the Odometry Prediction Stream. Details of the CNN encoder (with output $\text{FC}_{3}$) follows in \autoref{tab:cnne_details1}}
\label{tab:odo_details1}
\end{table}

\begin{table}[h]
\centering
\resizebox{\linewidth}{!}{
\begin{tabular}{ccccccc}
\toprule
Layer & Type & Filters & Size & Activation & Input & Output \\
\midrule
$\text{In}_4$ & Input & & & & & $\text{C}_1$ \\
\midrule
$\text{C}_1$ & Conv & 32 & 3$\times$3 & \emph{ReLU} & $\text{In}_2$ & $\text{C}_2$ \\
$\text{C}_2$ & Conv & 32 & 3$\times$3 & \emph{ReLU} & $\text{C}_1$ & $\text{P}_1$ \\
$\text{P}_1$ & MaxPool & & 2$\times$2 & & $\text{C}_2$ & $\text{C}_3$ \\
\midrule
$\text{C}_3$ & Conv & 64 & 3$\times$3 & \emph{ReLU} & $\text{P}_1$ & $\text{C}_4$ \\
$\text{C}_4$ & Conv & 64 & 3$\times$3 & \emph{ReLU} & $\text{C}_4$ & $\text{P}_2$ \\
$\text{P}_2$ & MaxPool & & 2$\times$2 & & $\text{C}_4$ & $\text{C}_5$ \\
\midrule
$\text{C}_5$ & Conv & 128 & 3$\times$3 & \emph{ReLU} & $\text{P}_2$ & $\text{C}_6$ \\
$\text{C}_6$ & Conv & 128 & 3$\times$3 & \emph{ReLU} & $\text{C}_5$ & $\text{P}_3$ \\
$\text{P}_3$ & MaxPool & & 2$\times$2 & & $\text{C}_6$ & $\text{C}_7$ \\
\midrule
$\text{C}_7$ & Conv & 256 & 3$\times$3 & \emph{ReLU} & $\text{P}_3$ & $\text{C}_8$ \\
$\text{C}_8$ & Conv & 256 & 3$\times$3 & \emph{ReLU} & $\text{C}_7$ & $\text{C}_8$ \\
$\text{P}_4$ & MaxPool & & 2$\times$2 & & $\text{C}_8$ & $\text{C}_9$ \\
\midrule
$\text{C}_9$ & Conv & 512 & 3$\times$3 & \emph{ReLU} & $\text{P}_4$ & $\text{C}_{10}$ \\
$\text{C}_{10}$ & Conv & 512 & 3$\times$3 & \emph{ReLU} & $\text{C}_9$ & $\text{P}_5$ \\
$\text{P}_5$ & MaxPool & & 2$\times$2 & & $\text{C}_{10}$ & $\text{FC}_1$ \\
\midrule
$\text{FC}_1$ & Dense & & 1024  & \emph{ReLU} & $\text{P}_5$ & $\text{FC}_2$ \\
$\text{FC}_2$ & Dense & & 256 & \emph{ReLU} & $\text{FC}_1$ & $\text{FC}_{3}$ \\
$\text{FC}_3$ & Dense & & 128 & \emph{tanh} & $\text{FC}_2$ & $\text{LSTM}_{dec2}$ \\
\bottomrule
\end{tabular}
}
\caption{Details of the CNN encoder used to condition the output of the Odometry prediction stream. Conv stands for 2D convolution, MaxPool stands for 2D max pooling and UpSample stands for 2D upsampling operations.}
\label{tab:cnne_details1}
\end{table}  

\section{Database Statistics}
\begin{figure}[h]
    \centering
    \includegraphics[width=\textwidth, height = 5cm,  keepaspectratio]{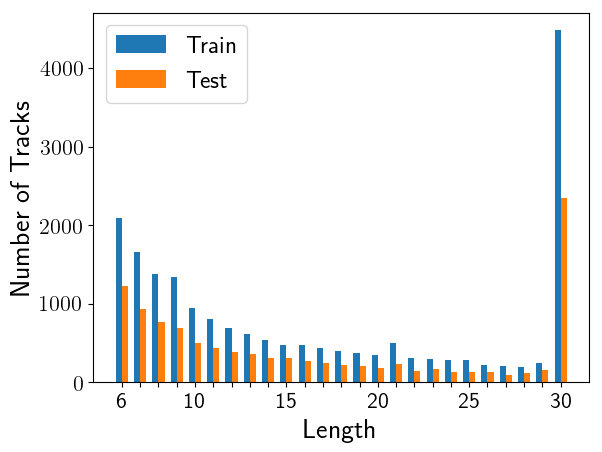}
    \caption{Length of recovered pedestrian tracks in Cityscapes.}
    \label{fig:trackstats}
\end{figure}
In \autoref{fig:trackstats} we plot the number of pedestrian tracks of lengths from 6 to 30. The track length distribution is consistent across training and test sets. We observe that there are many long tracks which stretch over the entire length (30) of the sequence.

\section{Evaluation with Varying Size of LSTM}
\begin{table}[h]
\begin{minipage}[b]{\linewidth}
\centering
\resizebox{\linewidth}{!}{
\begin{tabular}{ccccc}
\toprule
 Method & LSTM size & Odometry & MSE & $\mathcal{L}$ \\
\midrule
LSTM & 128 & None & 650 & 7.77 \\
LSTM & 512 & None & 705 & 8.15 \\ 
LSTM-Bayesian & 128  & None & \textbf{618} & \textbf{4.13} \\ 
LSTM-Bayesian & 512  & None & 619 & 4.16 \\ 
\bottomrule
\end{tabular}
}
\caption{Evaluation with varying size of LSTM ($|\text{B}_{\text{p}}|=8$).}
\label{tab:varlastmsize}
\end{minipage}
\end{table}
In the main paper, we evaluate all models constant LSTM vector size of 128. Here,  we report results for the (unconditioned) one stream homoscedastic LSTM encoder-decoder model and the one stream Bayesian LSTM encoder-decoder model using a vector size of 512 In \autoref{tab:varlastmsize}. We see that the homoscedastic version with 512 neurons performs worse than the version with 128 neurons. This is because the larger LSTM over-fits to the bounding box estimation noise in dataset. However, the Bayesian versions have comparable performance, due to dropout which prevents overfitting. 

\begin{figure*}[!t]
\begin{tabular}{ccc}

  \includegraphics[width=0.32\linewidth]{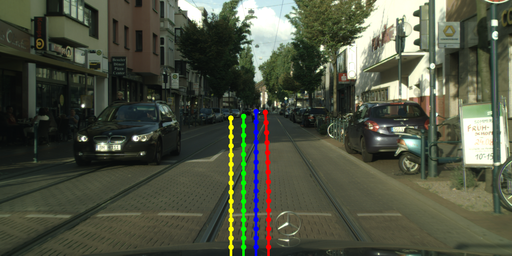} &

  \includegraphics[width=0.32\linewidth]{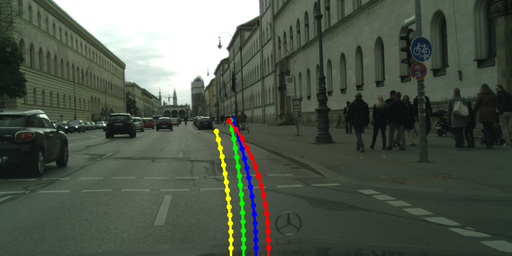} &

  \includegraphics[width=0.32\linewidth]{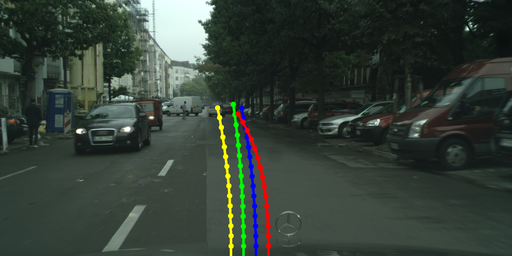} \\
  
  \includegraphics[width=0.32\linewidth]{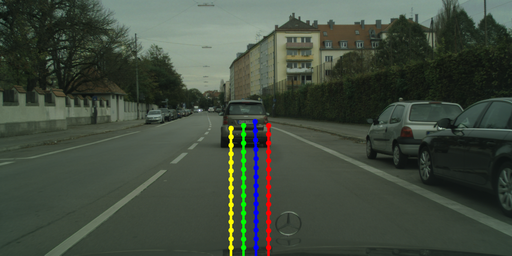} &

  \includegraphics[width=0.32\linewidth]{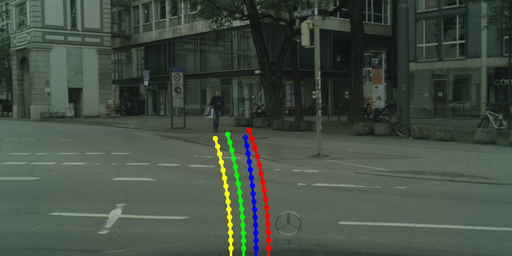} &

  \includegraphics[width=0.32\linewidth]{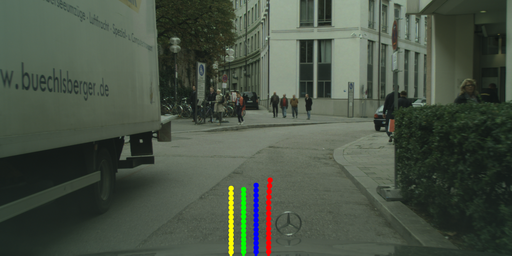} \\

  \includegraphics[width=0.32\linewidth]{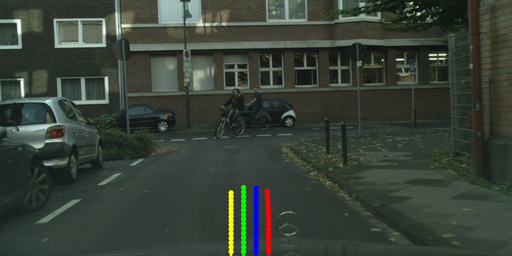} &

  \includegraphics[width=0.32\linewidth]{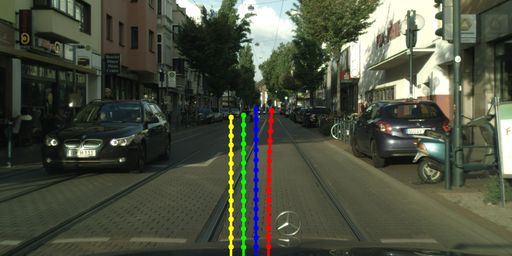} &

  \includegraphics[width=0.32\linewidth]{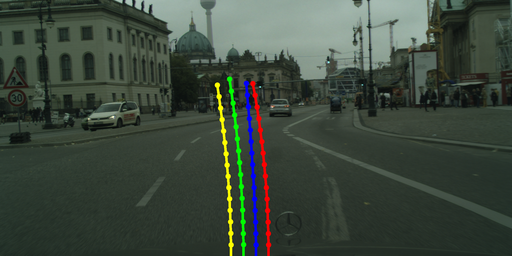} \\

  \includegraphics[width=0.32\linewidth]{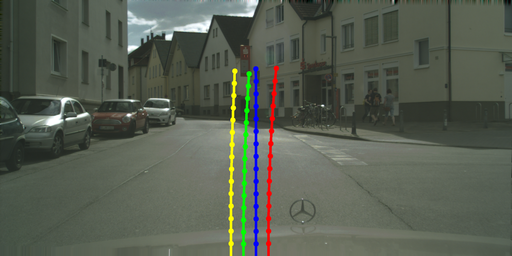} &

  \includegraphics[width=0.32\linewidth]{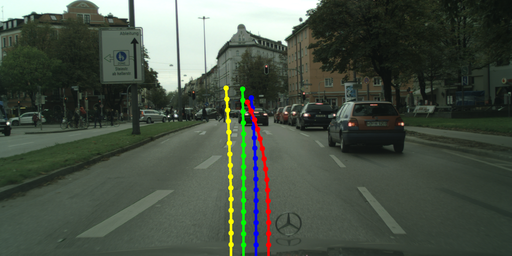} &

  \includegraphics[width=0.32\linewidth]{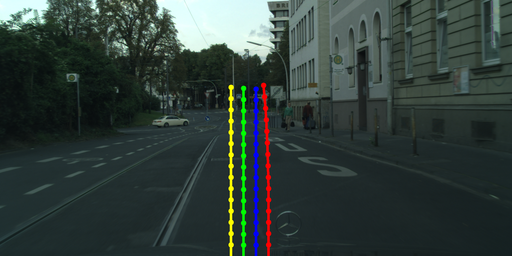} \\
\end{tabular}
\caption{Odometry prediction: We show predicted odometry for 15 time-steps as points (bottom to top) over-layed on the last visual observation. The distance and angle between subsequent points is the predicted (proportional) speed and steering angle. Color codes: \textcolor{blue}{Blue}: Ground-truth, \textcolor{red}{Red}: Kalman Filter, \textcolor{yellow}{Yellow}: Our LSTM without visual input, \textcolor{green}{Green}: Our LSTM with visual input.} \label{fig:steervex}
\end{figure*}

\section{Visualization of Odometry Prediction}
Visual examples of odometry prediction in \autoref{fig:steervex}.

\section{Additional Evaluation of our Two-stream Model}
\begin{table}[h]
\centering
\begin{tabular}{ccccc}
\toprule
 Method & Streams  & Visual & MSE & $\mathcal{L}$\\
\midrule
LSTM & Two & RGB & 516 & 5.15\\
LSTM-Aleatoric & Two & RGB & 618 & 4.92\\
LSTM-Bayesian & Two & RGB & \textbf{505} & \textbf{3.92}\\
\bottomrule
\end{tabular}
\caption{Evaluation of Two-stream models ($|\text{B}_{\text{p}}|,|\text{O}_{\text{p}}| = 8$).}
\label{tab:bothstreameval2}
\end{table}

Here, we compare our Bayesian Two-stream model (Figure 2, of main paper) to, 
\begin{enumerate*} 
    \item A homoscedastic Two-stream LSTM encoder-decoder model (LSTM).
    \item A heteroscedastic Two-stream LSTM encoder-decoder (LSTM-Aleatoric).
\end{enumerate*}
Note that, both models have the same odometry prediction stream as our Bayesian Two-stream LSTM model (LSTM-Bayesian). The results mirror the evaluation of only the bounding box prediction stream. We see that the heteroscedastic LSTM (LSTM-Aleatoric, 2nd row) outperforms the homoscedastic LSTM (2nd row) with respect to the $\operatorname{\mathcal{L}}$ metric. This means that the heteroscedastic Two-stream LSTM learns to capture uncertainty and assigns higher probability to the true bounding box sequence. However, when epistemic uncertainty is not modelled, aleatoric uncertainty tried to compensate and this leads to poorer MSE. Finally, our Bayesian Two-stream LSTM (3rd row) outperforms all other methods.

\section{Additional Analysis of the Quality of our Uncertainty Metric}

\begin{figure}[h]

\resizebox{0.49\textwidth}{!}{
\begin{tabular}{cc}

    \includegraphics[width=0.245\textwidth, height=0.16\textheight]{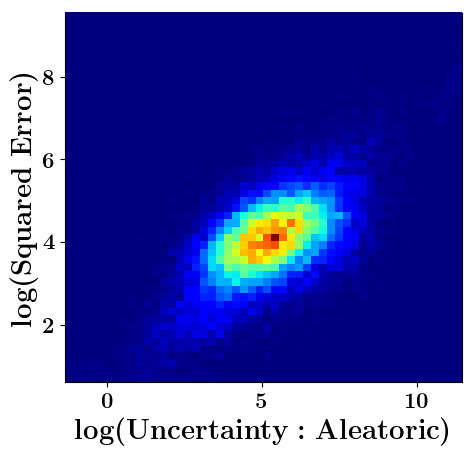} &
    \includegraphics[width=0.245\textwidth, height=0.16\textheight]{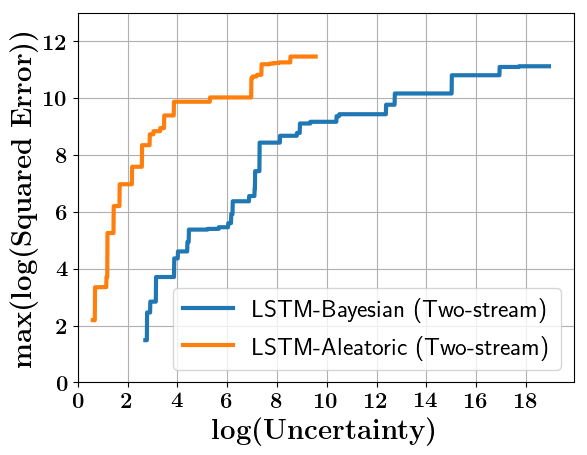}\\

\end{tabular}
}
\caption{Plot 1 - uncertainty versus squared error, plot 3 - uncertainty versus \emph{maximum} observed squared error.}\label{fig:uncertinitymetric2}
\end{figure}

We compare the quality of the uncertainty metric obtained with our Two-stream LSTM-Bayesian model (Figure 3, of main paper) to that of the Two-stream LSTM-Aleatoric (the heteroscedastic Two-stream LSTM encoder-decoder in the previous section, which models only aleatoric uncertainty). In plot 1 of \autoref{fig:uncertinitymetric2} the aleatoric uncertainty to the $\log$ squared error of the mean of the predictive distribution of the Two-stream LSTM-Aleatoric model is shown. We see that the distribution is more spread-out with more outliers compared to our Two-stream LSTM-Bayesian model (plot 1, Figure 3, of main paper). In plot 2 of \autoref{fig:uncertinitymetric2} the \emph{maximum} $\log$ squared error (of the mean of the predictive distribution) observed at a certain predicted uncertainty in the test test is shown for both our Two-stream Bayesian model and Two-stream LSTM-Aleatoric. We see that the correlation is poor compared to our Two-stream LSTM-Bayesian model (also in plot 3, Figure 3, of main paper). In particular, the maximum observed $\log$ squared error rises very sharply. Therefore, for a robust error bound it is essential to model both epistemic and aleatoric uncertainty.

\section{Additional Video Results}
We include video results of prediction in \textbf{video.mp4}. We include examples of both point estimates and predictive distributions. We include point estimates for comparison against the Kalman Filter and One-stream baselines. The examples show accurate prediction by our Two-stream model over 15 time-steps into the future.

\end{appendix}

\end{document}